\documentclass[10pt,twocolumn,letterpaper]{article}

\usepackage{iccv}
\usepackage{times}
\usepackage{epsfig}
\usepackage{graphicx}
\usepackage{amsmath}
\usepackage{amssymb}
\usepackage{subfigure}
\usepackage{float}

\usepackage{booktabs}
\usepackage{epstopdf}
\usepackage{threeparttable}
\usepackage{multirow}

\usepackage[breaklinks=true,bookmarks=false]{hyperref}

\iccvfinalcopy 

\def\iccvPaperID{****} 

\ificcvfinal\fi

\begin{document}

\title{ReSup: Reliable Label Noise Suppression for Facial Expression Recognition}

\author{Xiang~Zhang$^1$,
Yan~Lu$^2$,
Huan~Yan$^1$,
Jingyang~Huang$^1$,
Yusheng Ji$^3$,
Yu~Gu$^4$\\
$^1$ Hefei University of Technology, $^2$ University of Sydney\\
$^3$ National Institute of Informatics, $^4$ University of Electronic Science and Technology of China\\
}
\maketitle
\ificcvfinal\thispagestyle{empty}\fi

\begin{abstract}

%
%
%
Because of the ambiguous and subjective property of the facial expression recognition (FER) task, the label noise is widely existing in the FER dataset. For this problem, in the training phase, current FER methods often directly predict whether the label of the input image is noised or not, aiming to reduce the contribution of the noised data in training. However, we argue that this kind of method suffers from the low reliability of such noise data decision operation. It makes that some mistakenly abounded clean data are not utilized sufficiently and some mistakenly kept noised data disturbing the model learning process. In this paper, we propose a more reliable noise-label suppression method called ReSup (\textbf{Re}liable label noise \textbf{Sup}pression for FER). First, instead of directly predicting noised or not, ReSup makes the noise data decision by modeling the distribution of noise and clean labels simultaneously according to the disagreement between the prediction and the target. Specifically, to achieve optimal distribution modeling, ReSup models the similarity distribution of all samples. To further enhance the reliability of our noise decision results, ReSup uses two networks to jointly achieve noise suppression. Specifically, ReSup utilize the property that two networks are less likely to make the same mistakes, making two networks swap decisions and tending to trust decisions with high agreement. Extensive experiments on three popular benchmarks show that the proposed method significantly outperforms state-of-the-art noisy label FER methods by 3.01\% on FERPlus becnmarks. Code:\href{URL}{https://github.com/purpleleaves007/FERDenoise}

\end{abstract}

\section{Introduction}
\label{sec:intro}
  

\begin{figure}[ht]
  \centering
   \includegraphics[width=0.9\linewidth]{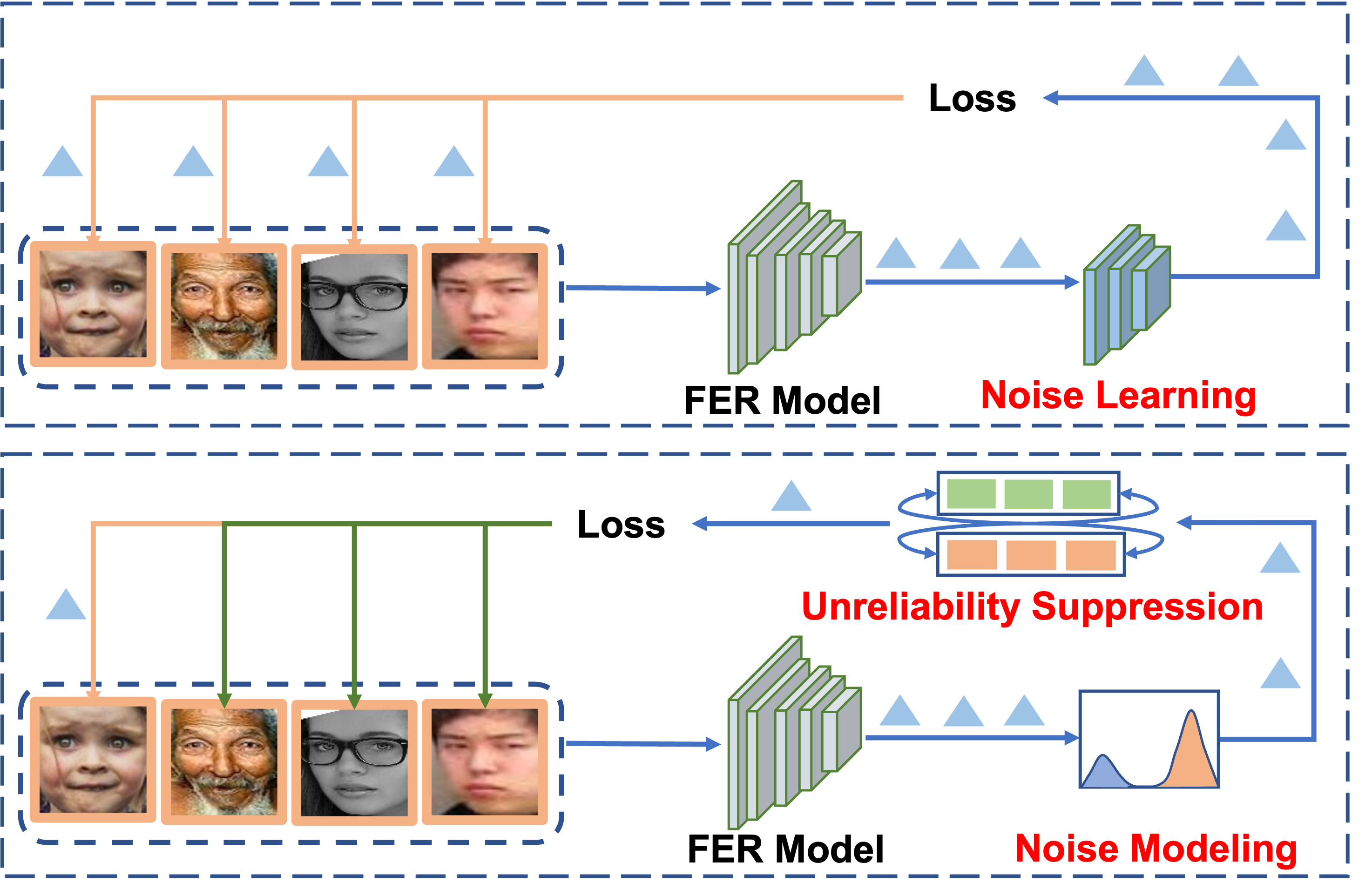}
   \caption{Comparison of label noise suppression process of different noisy label FER methods. The top is the current schemes and the bottom is ReSuP. ReSuP generates more reliable weights by noise modeling and suppresses unreliable weights by unreliability suppression design in addition to suppressing noisy labels. Triangles represent unreliable weights.}
   \label{fig:int}
   \vspace{-0.2in}
\end{figure}


Facial expression recognition (FER) has become a crucial service in various real-world applications, such as healthcare~\cite{lisetti2003developing}, surveillance~\cite{al2012facial}, and virtual reality~\cite{lou2019realistic}. 
%
It aims to recognize specific human emotions from the given facial images.
%
However, for the large-scale FER dataset collected from the Internet, it is difficult to achieve high-quality annotations due to the subjectivity of annotators and the ambiguity of facial expressions, and these low-quality annotations form label noises.
Therefore, how to suppress label noises in FER tasks has become a research hotspot and attracted more and more attention\cite{wang2020suppressing, she2021dive, jiang2021boosting, zhang2021relative, zeng2018facial, yan2022mitigating}. 



For this challenge, existing FER methods typically incorporate an importance learning branch to estimate the importance weight of each image, which determines whether the label of the input image is noisy or not
~\cite{wang2020suppressing,she2021dive,zhang2021relative}. However, we argue that these methods suffer from the low reliability of such noise decision operation. 
Such operation usually generates unreliable weights and makes that some mistakenly abounded clean data are not utilized sufficiently and some mistakenly kept noised data disturbing the model learning process.
These unreliable weights originate from two perspectives: the noise decision process and the FER model itself. The former unreliable weights are due to overfitting of the importance learning branch as a result of the strong learning ability of deep neural networks (DNNs)~\cite{zhang2021relative}. Furthermore, such noise decision process only considers information from a single sample~\cite{wang2020suppressing} a batch~\cite{she2021dive}, neglecting global information and resulting in unreliable decision making. In addition to the unreliable weights caused by the noise decision process, the FER model itself unavoidably produces some unreliable outputs (the inputs of the importance learning branch), leading to further unreliable weights. These unreliable weights accumulate during the entire learning process and affect current and future learning stages. Unfortunately, existing methods do not address \textbf{how to mitigate the effects of these unreliable weights.} Novel methods are required to address these limitations and improve the reliability and accuracy of noisy label FER.

In this paper, we present a novel approach called ReSup. The main objective of ReSup is to suppress noisy labels and unreliable weights, as illustrated in Figure \ref{fig:int}. Specifically, instead of directly predicting noised or not, ReSup makes the noise decision by modeling he joint distribution of noise and clean labels.
This is motivated by the memorization effect of deep neural networks (DNNs), where the model tends to memorize correctly labeled samples first~\cite{zhang2021understanding, arpit2017closer, han2018co}, leading to noisy samples having higher loss during early epochs of training~\cite{arazo2019unsupervised, zhang2019identity}.
But, to achieve optimal distribution modeling, ReSup propose to model the similarity (cosine similarity of predictions and targets) distribution of all samples rather than the loss, which reduces the confusion between noisy and clean distributions. The fitted noise model is then used to provide importance weights for each sample based on its similarity, without using neural network branches to avoid overfitting. In addition, the proposed scheme can take into account the global distribution of all samples. 
Furthermore, ReSup mitigate the effect of the unreliable weights by leveraging the agreement maximization principles~\cite{blum1998combining, sindhwani2005co}, which suggest that two different networks would agree on most samples except for noisy samples~\cite{wei2020combating} and thus can filter different types of errors. 
Inspired by the agreement maximization principles, ReSup employs two different networks to provide importance weights to each other, to prevent the accumulation of errors caused by unreliable weights. We also introduce a consistency loss that assigns large losses to samples with small agreement to prevent the model from fitting samples with unreliable weights, since the samples with small agreements usually are noisy samples. In summary, our contributions include:

\begin{itemize}
    \item To avoid extra unreliable weights caused by the DNN-based importance learning branch, a novel label noise modeling method based on similarity distribution statistics is proposed to estimate the importance weights.
    \item We propose ReSup to suppress label noise in FER. ReSup satisfactorily mitigates the effect of the unreliable weights by leveraging a weight exchange strategy and a consistency loss.
    \item Extensive experiment results demonstrate that the proposed ReSup significantly outperforms state-of-the-art noisy label FER solutions on multiple FER benchmarks with different levels of label noise.
\end{itemize}

\begin{figure*}[ht]
  \centering
   \includegraphics[width=0.75\linewidth]{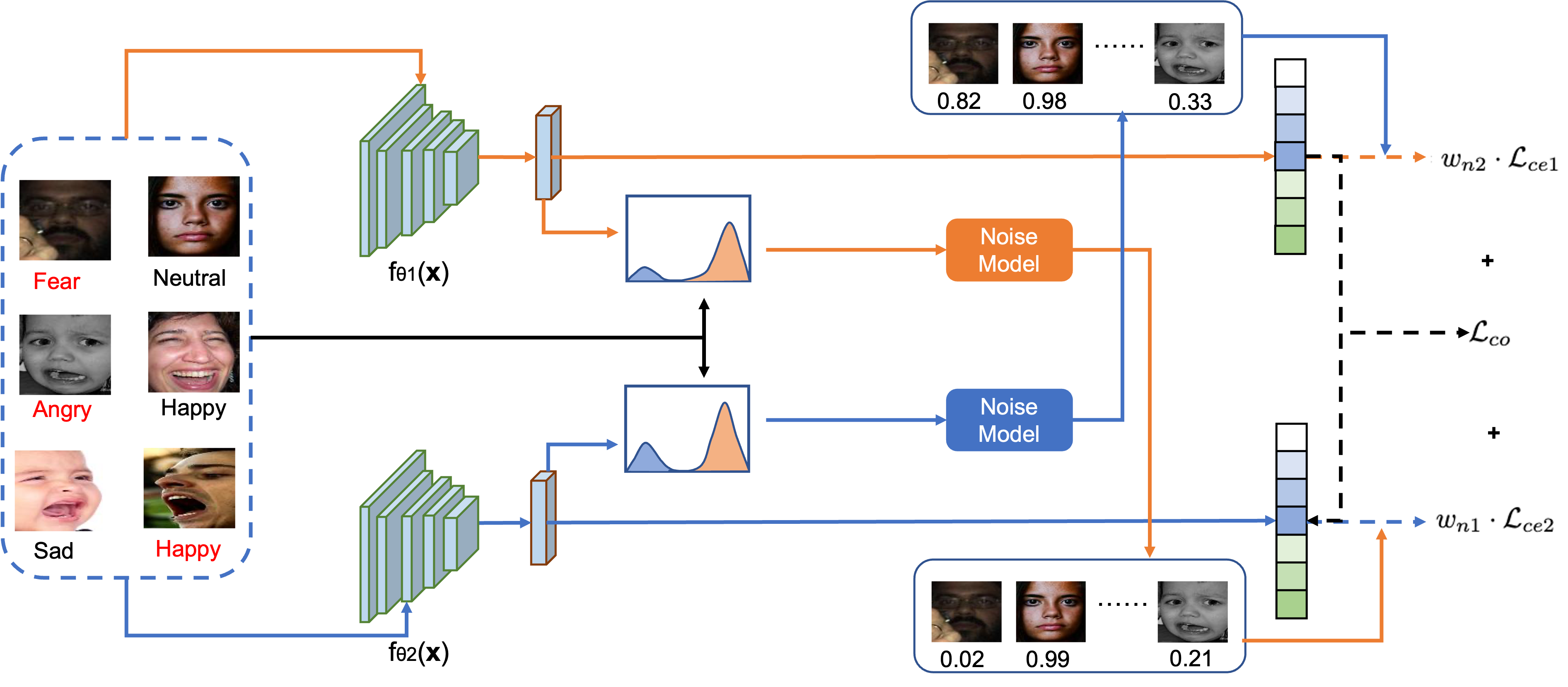}
   \caption{The ReSup framework.}
   \label{fig:onecol}
      \vspace{-0.2in}
\end{figure*}

\section{Related Work}
\label{sec:relatedwork}

\subsection{Facial Expression Recognition}
The categorization of FER methods can be broadly divided into two groups based on the features used: handcraft-based and learning-based approaches. Earlier research mainly relied on handcrafted features~\cite{bazzo2004recognizing,dalal2005histograms}, which capture the folds and geometry changes caused by facial expressions~\cite{zhong2012learning,khan2017co}. Nevertheless, researchers have gradually exposed the limitations of handcraft-based methods, particularly in-the-wild scenarios. Fortunately, with the advancement in computational ability and the rapid development of large-scale datasets, e.g., AffectNet~\cite{mollahosseini2017affectnet}, RAF-DB~\cite{li2017reliable}, and EmotioNet~\cite{fabian2016emotionet}, recent studies mainly focus on deep learning~\cite{gu20213} as a better alternative. Ruan \etal.~\cite{ruan2021feature, ruan2022adaptive} proposed the feature decomposition method and the reconstruction learning algorithm for effective FER. RAN~\cite{wang2020region} proposed a region attention network to overcome the pose and occlusion challenges in FER. Transformers~\cite{xue2021transfer} have also been introduced into the FER due to its powerful global information awareness. However, the presence of label noise in large-scale datasets remains a significant challenge for FER in in-the-wild scenarios.

In recent years, several algorithms are proposed to address the label noise challenge in FER. Among these, SCN~\cite{wang2020suppressing}, RUL~\cite{zhang2021relative}, and DMUE~\cite{she2021dive} use neural network branches to estimate importance weight for each sample, in order to predict whether the label of the input image is noised or not. However, the strong learning capability of DNNs can lead to overfitting of the importance learning branch and the fact that these approaches do not exploit global information, both leading to unreliable weights beyond the FER model itself. To avoid this issue, PT~\cite{jiang2021boosting} selected samples with small losses as clean samples for training, without relying on a neural network branch to learn importance weights. However, PT requires setting a threshold based on the exact noise level and dataset used, which is impractical in real-world scenarios. In contrast, our proposed scheme uses unsupervised noise modeling to mitigate these issues without requiring prior knowledge, and suppresses unreliable weights by unreliability suppression design in addition to suppressing noisy labels.


\subsection{Noisy Label Learning}

Learning with noisy labels has been extensively studied in the computer vision community, and current approaches can be broadly categorized into four following groups~\cite{song2022learning}: robust architecture, robust regularization, robust loss function, and robust data.
Robust architecture-based methods usually added a noise adaptation layer~\cite{barsoum2016training} at the top of the network~\cite{srivastava2014dropout} to learn label transition proces or designed a noise-tolerant architecture~\cite{xiao2015learning} to reliably support more diverse types of label noise. 
Robust regularization-based methods aimed to explicitly~\cite{wei2021open} or implicitly~\cite{lukasik2020does} induce DNNs to be less likely to overfit the noisy labels. By avoid overfitting in training, the robustness to label noise improves
with regularization techniques such as data-augmentation~\cite{shorten2019survey} and weight decay~\cite{krogh1991simple}.
Robust loss function-based methods seeked to design a loss function that is robust to noisy labels, to prevent DNNs from fitting to the noisy samples~\cite{ghosh2017robust,liu2020peer}. 
Robust data-based methods aimed to select clean samples from noisy data by threshold~\cite{han2018co} or weight~\cite{arazo2019unsupervised}. Threshold-based algorithms, which require an artificial threshold to select clean samples and discard noisy samples, have been explored in classical robust data-based methods~\cite{han2018co, malach2017decoupling, yu2019does}. Discarding noisy samples, however, potentially mistakenly discard clean samples and removes useful information about the data distribution \cite{wang2018iterative, arazo2019unsupervised}. To avoid this problem, recent approaches propose the use of two-component beta mixture models (BMM) ~\cite{arazo2019unsupervised} or Gaussian mixture models (GMM) ~\cite{li2020dividemix} that rely on losses to model the label noise and assign weights to the samples based on the fitted noise model to suppress the noisy labels. Based on the memory effect mentioned in section \ref{sec:intro}, these schemes believe that the loss distributions of clean and noisy samples are significantly different. However, these state-of-the-art schemes rely on loss distribution for noise modeling, which may not be effective in noisy label FER due to the inter-class similarity of facial expressions. We propose to model label noise by relying on the similarity and demonstrate its effectiveness in section \ref{subsec:abl}.

\section{Method}

\subsection{Overview of ReSup}

FER can be formulated as the problem of learning a model $f_\theta(\boldsymbol{x})$ from a set of facial images $T = \left \{ (\boldsymbol{x}_i,\boldsymbol{y}_i) \right \} _{i=1}^{N} $ with $\boldsymbol{y}_i \in \left \{ 0,1 \right \} ^{C} $ is the one-hot ground-truth label corresponding to $\boldsymbol{x}_i$. 

In our case, $f_\theta$ is a CNN and $\theta$ is the model parameters. Besides, the label $\boldsymbol{y}_i$ could differ from the unknown true label. During the learning, the parameters of the model are fitted by optimizing the cross-entropy (CE) loss:
\begin{equation}
    \mathcal{L}_{ce} = \sum_{i=1}^{N} \mathcal{L}_i = -\sum_{i=1}^{N} \boldsymbol{y}_i^{T} log(f_\theta (\boldsymbol{x}_i))
\end{equation}
where $f_\theta(\boldsymbol{x}_i)$ and $\mathcal{L}_i$ represent the softmax output and the loss produced by the model, respectively. For simplicity, we use $\boldsymbol{f}_i$ to represent $f_\theta(\boldsymbol{x}_i)$ in the remainder of the paper. The goal is to minimize the CE loss to obtain the optimal model parameters $\theta$. However, noisy labels can negatively impact the training process and lead to poor performance. Therefore, our goal is to mitigate the impact of label noise during the training process.

The framework of ReSup is depicted in Figure \ref{fig:onecol}. The proposed method comprises two components, namely label noise modeling and noise-robust learning. Label noise modeling aims to model the distribution of the label noise. The fitted noise model can provide reliable importance weights for each sample according to its similarity. These assigned weights are then used by noise-robust learning to effectively learn informative representations in the presence of label noise. To better achieve this goal, noise-robust learning uses a weight exchange strategy and a consistency loss to mitigate the effect of unreliable weights.

\subsection{Label Noise Modeling}


To enable a noise-robust learning approach (section \ref{subsec:crossde}), it is necessary to identify noisy samples in the training set $T$ first. The essential observation is simple: learning noisy labels takes longer than clean labels~\cite{arpit2017closer}, resulting in lower similarity for noisy samples (as depicted in Figure \ref{fig:short-c}). This distinction in similarity distributions allows for the differentiation between clean and noisy samples.

The similarity $\mathcal{S}_i$ of $\boldsymbol{x}_i$ is calculated as follows: 
\begin{equation}
    \mathcal{S}_i=\frac{\textstyle \sum_{k=1}^{c} \boldsymbol{y}_{ik} \cdot \boldsymbol{f}_{ik} }{\sqrt{\sum_{k=1}^{c}(\boldsymbol{f}_{ik})^{2} }}=cos(\boldsymbol{f}_i,\boldsymbol{y}_i)
\end{equation}
where $c$ is the number of classes.

Following previous studies~\cite{arazo2019unsupervised,li2020dividemix}, we first model the similarity distribution using a mixture model, and then use the generated noise model to provide a probability for each sample that it belongs to clean samples. The probability density function (pdf) of a mixture model of $K$ components on the similarity $\mathcal{S}$ can be defined as:
\begin{equation}
    p(\mathcal{S}) = \sum_{m=1}^{K}\delta _mp(\mathcal{S}|m) 
    \label{equ:mpd}
\end{equation}
where $\delta_m$ are the mixing coefficients of each pdf $p(\mathcal{S}|m)$. In our case, we fit a two-components (i.e. $K = 2$) mixture model to model the distribution of clean and noisy samples. 

We choose the more flexible BMM to model the similarity distribution. The pdf of the beta distribution is:
\begin{equation}
    p(\mathcal{S}|\alpha_m ,\beta_m ) = \frac{\Gamma (\alpha_m +\beta_m )}{\Gamma (\alpha_m)\Gamma (\beta_m )} \mathcal{S}^{\alpha_m -1}(1-\mathcal{S})^{\beta_m -1}  
\end{equation}
where $\Gamma (\cdot)$ is the Gamma function, $\mathcal{S}$ is the similarity, $\alpha_m, \beta_m > 0$, and the mixture pdf is given by substituting the above into equation \ref{equ:mpd}. 

We use an Expectation Maximization (EM) procedure to fit the BMM to the similarity of all samples in this study. After the noise model is fitted, we can obtain the probability of a sample belongs to clean samples (importance weights) through the posterior probability:
\begin{equation}
    w_{n} = p(m|\mathcal{S})=\frac{p(m)p(\mathcal{S}|m)}{p(\mathcal{S})}
\end{equation}
where $m= 1 (0)$ denotes clean (noisy) classes. $p(\mathcal{S}|m)$ is defined to be the posterior probability of the similarity $\mathcal{S}$ having been generated by component $m$. Why we model label noise by relying on the similarity but not loss~\cite{arazo2019unsupervised,li2020dividemix} are illustrated in section \ref{subsec:abl} and the supplemental material.

\subsection{Noise-Robust Learning}
\label{subsec:crossde}

Label noise modeling generates a noise model to assign importance weights to each training sample in $T$, and noise-robust learning aims to suppress the label noise in $T$ and extract meaningful knowledge by leveraging these importance weights. To better achieve this goal, we propose a weight exchange strategy along with a consistency loss to eliminate the effect caused by unreliable weights. The proposed method consists of two DNNs denoted by $f_{\theta1}(\boldsymbol{x})$ and $f_{\theta2}(\boldsymbol{x})$.



\textbf{Network.} For ReSup, both $f_{\theta1}(\boldsymbol{x})$ and $f_{\theta2}(\boldsymbol{x})$ can be used to predict facial expression alone, but during the training stage, the parameters of the two networks are updated simultaneously by a joint loss. Specifically, the joint loss function $\mathcal{L}_{jo}$ is constructed as follows:
\begin{equation}
\label{equ:ljo}
    \mathcal{L}_{jo} =  \mathcal{L}_{wc} + \lambda \mathcal{L}_{co}
\end{equation}

In the joint loss function, the first part $\mathcal{L}_{wc}$ is the weighted CE loss of the two networks, which reliably uses the importance weights by relying on the weight exchange strategy to suppress label noise and learn useful knowledge. The second part $\mathcal{L}_{co}$ is the consistency loss, which is used to further attenuate the effect of unreliable weights. $\lambda$ is the hyperparameter to control the influence of $\mathcal{L}_{co}$.


\textbf{Weight Exchange Strategy Based Weighted CE Loss.} The importance weights are utilized to weigh the CE loss for suppress label noise. To avoid errors caused by unreliable weights being accumulated, the weights of $f_{\theta1}(\boldsymbol{x})$ and $f_{\theta2}(\boldsymbol{x})$ are exchanged. Intuitively, different networks have different learning abilities and therefore can filter out different types of errors~\cite{han2018co,yu2019does}, thus in the weight exchange procedure, the error flows can be reduced by peering networks mutually. The weighted CE loss is formulated as:
\begin{equation}
    \mathcal{L}_{wc}= w_{n2}\cdot \mathcal{L}_{ce1} + w_{n1}\cdot \mathcal{L}_{ce2}
\end{equation}
where $\mathcal{L}_{ce1}$ and $\mathcal{L}_{ce2}$ represents the CE loss of  $f_{\theta1}(\boldsymbol{x})$ and $f_{\theta2}(\boldsymbol{x})$, respectively. $w_{n1}$ and $w_{n2}$ denotes the importance weight from $f_{\theta1}(\boldsymbol{x})$ and $f_{\theta2}(\boldsymbol{x})$, respectively.

\textbf{Consistency Loss.} From the view of agreement maximization principles~\cite{blum1998combining,sindhwani2005co}, different networks are unlikely to agree on noisy labels, which means that the probability of two networks making the same unreliable decisions for the same noisy samples simultaneously is small. Thus, we utilize a consistency loss to evaluate the agreement between the two networks. The consistency loss imposes high loss values on samples with low agreement to further discourage the model from fitting samples with unreliable weights. When there is a small agreement, the probability that the sample belongs to noise is high. A high consistency loss makes the model more likely to fit the prediction of another network instead of the intended target. The consistency loss is as follows:
\begin{equation}
    \mathcal{L}_{co}= \frac{1}{c} \sum_{i=1}^{N} \sum_{k=1}^{c}(\boldsymbol{f}_{ik1}-\boldsymbol{f}_{ik2})^{2} 
\end{equation}
$\boldsymbol{f}_{ik1}$ and $\boldsymbol{f}_{ik2}$ denotes the outputs of the “softmax" layer in $f_{\theta1}(\boldsymbol{x})$ and $f_{\theta2}(\boldsymbol{x})$ for sample $x_i$ on class $k$, respectively. 


The closest approach to our model is JoCoR~\cite{wei2020combating}, and JoCoR also consists of two networks and a contrastive loss. However, JoCoR uses the average loss of the two networks to select clean samples based on a threshold. Unlike JoCoR, ReSup uses two networks to provide importance weights to each other and employs a consistency loss to eliminate the effect of unreliable weights. We have also implemented a JoCoR-based scheme in our experiments.

\begin{figure}
  \centering
    \includegraphics[width=1\linewidth]{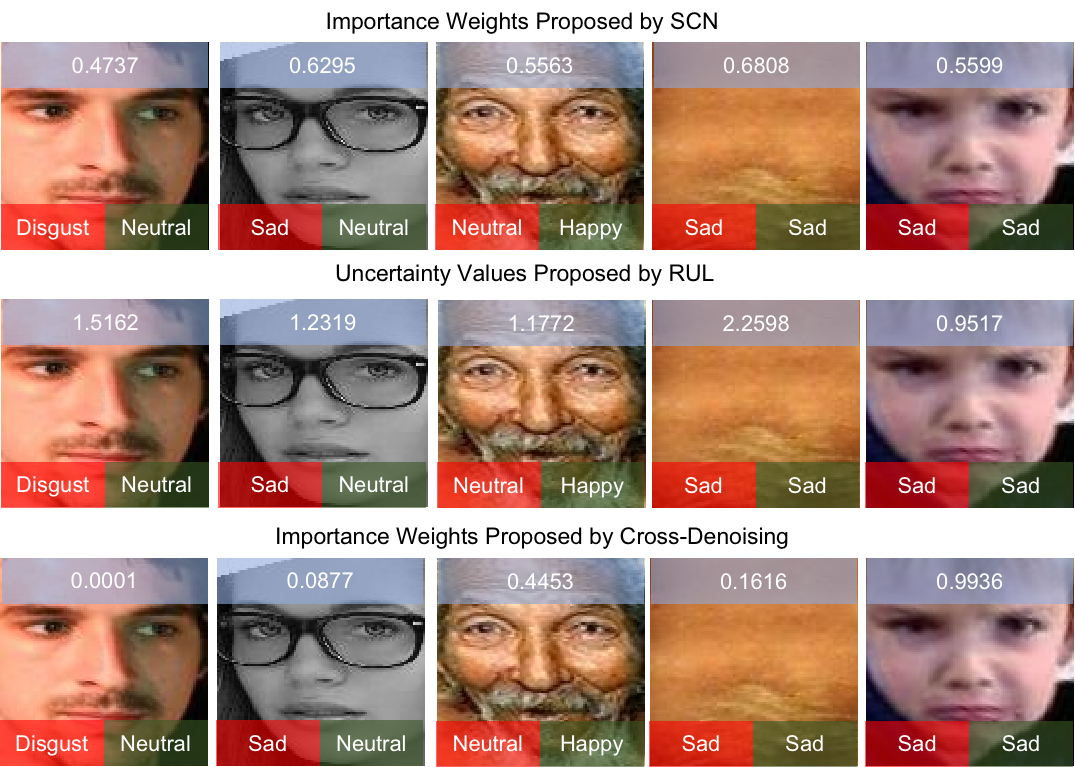}
    \caption{Visualization of the confidence score of SCN, RUL, and ReSup. All models are trained on RAF-DB with 30\% noise. The importance weights/uncertainty values are marked at the top. The true labels and the labels for training are marked in the lower right and left corners, respectively. For noisy samples, the uncertainty values should be large but the importance weights should be small.}
    \label{fig:weiexa}
\end{figure}

\begin{figure}
  \centering
    \includegraphics[width=1\linewidth]{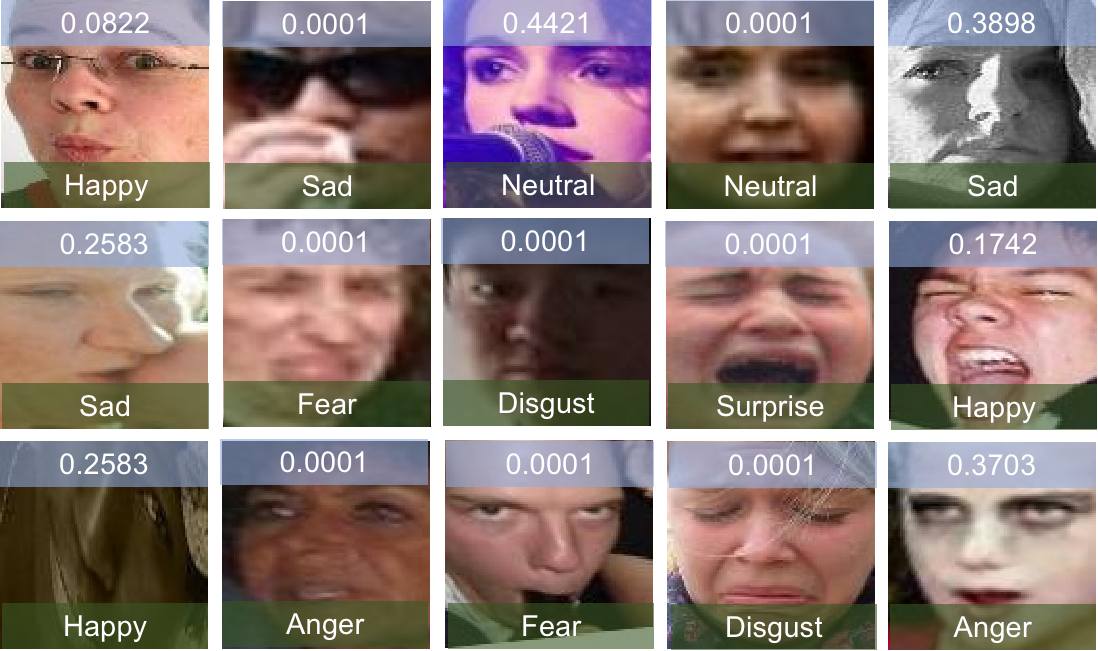}
    \caption{Some examples of RAF-DB (w/o synthetic noisy labels) with low importance weights. The importance weights are marked at the top, and the true labels are marked at the bottom.}
    \label{fig:weiraf}
\end{figure}

\begin{table}
  \centering
  \caption{Comparison with other state-of-the-art noisy label FER schemes. Results are computed as the mean of the last 5 epochs.}
  \begin{tabular}{@{}lcccc@{}}
    \toprule
    Method & Noise & RAF-DB & FERPlus & AffectNet \\
    \midrule
    SCN~\cite{wang2020suppressing} & 10\% & 82.18\% & 84.28\% & 61.57\% \\
    DMUE~\cite{she2021dive} & 10\% & 83.19\% & - & - \\
    RUL~\cite{zhang2021relative} & 10\% & 86.22\% & 86.93\% & 62.89\%\\
    PT~\cite{jiang2021boosting} & 10\% & 87.28\% & 85.04\% & -\\
    Ours & 10\% & \textbf{88.43\%} & \textbf{87.82\%} & \textbf{64.29\%}\\
    \midrule
    SCN~\cite{wang2020suppressing} & 20\% & 80.10\% & 83.17\% & 60.83\% \\
    DMUE~\cite{she2021dive} & 20\% & 81.02\% & - & - \\
    RUL~\cite{zhang2021relative} & 20\% & 84.34\% & 85.05\% & 61.74\%\\
    PT~\cite{jiang2021boosting} & 20\% & 86.25\% & 84.27\% & -\\
    Ours & 20\% & \textbf{87.29\%} & \textbf{87.08\%} & \textbf{63.97\%}\\
    \midrule
    SCN~\cite{wang2020suppressing} & 30\% & 77.46\% & 82.47\% & 58.80\% \\
    DMUE~\cite{she2021dive} & 30\% & 79.41\% & - & - \\
    RUL~\cite{zhang2021relative} & 30\% & 82.06\% & 83.90\% & 60.77\%\\
    PT~\cite{jiang2021boosting} & 30\% & 84.32\% & 83.73\% & -\\
    Ours & 30\% & \textbf{86.86\%} & \textbf{86.74\%} & \textbf{62.89\%}\\
    \bottomrule
  \end{tabular}
  \label{tab:noisy}
  \vspace{-0.1in}
\end{table}

\section{Experiments}

In this section, we evaluate the effectiveness of ReSup on synthetic label noise datasets and in-the-wild benchmarks.


\subsection{Datasets}

\textbf{RAF-DB}~\cite{li2017reliable} is the first in-the-wild dataset containing basic or compound expressions, including nearly 30k facial images annotated with basic or compound expressions by 40 trained human annotators. In our experiments, we only use images of six basic expressions (happy, surprise, sad, anger, disgust, fear) as well as neutral, which lead to 12,271 images for training and 3,068 images for testing.

\textbf{FERPlus}~\cite{barsoum2016training} is an extension of FER2013, which is used in the ICML 2013 Challenge. It is a large-scale dataset collected through the Google search engine. It contains 28,709 training images, 3,589 validation images, and 3,589 test images, which are resized to 48×48 pixels. FERPlus includes eight emotional states (six basic emotions, neutral and contempt), and each image is labeled by ten human annotators.

\textbf{AffectNet}~\cite{mollahosseini2017affectnet} is the largest in-the-wild facial expression dataset by far. It contains nearly 450K manually annotated facial images collected from the Internet by three major search engines with emotion-related keywords. This dataset has an imbalanced training set and a balanced validation set. Following previous work~\cite{zeng2018facial,zhao2021robust,chen2020label}, we selected approximately 280,000 and 3,500 images for training and testing, the classes are the same as RAF-DB.

\subsection{Implementation Details}

In our experiments, we adopt ResNet-18~\cite{he2016deep} pretrained on the MS-Celeb-1M~\cite{guo2016ms} as $f_{\theta1}(\boldsymbol{x})$ and ResNet-18 pretrained on the ImageNet~\cite{deng2009imagenet} as $f_{\theta2}(\boldsymbol{x})$ for fair comparisons with previous works~\cite{wang2020suppressing,zhang2021relative,jiang2021boosting}. To ensure consistency of our results, we only use the outputs of $f_{\theta1}(\boldsymbol{x})$ during testing and discard $f_{\theta2}(\boldsymbol{x})$.  Images we used are aligned and cropped with three landmarks, then resized to $224\times 224$ pixels, and augmented by random horizontal flipping, random erasing, and random cropping. During training, we use a batch size of 96 and Adam as the optimizer with an initial learning rate of $0.0002$. We divide the learning rate by $10$ at epoch $10$ and $20$ for RAF-DB and FERPlus, and at epoch $5$ and $10$ for AffectNet. Training concludes at epoch $30$ for RAF-DB and FERPlus, and at epoch $20$ for AffectNet. We set the hyperparameter $\lambda$ to $5$ by default based on our ablation studies. Our implementation is based on the Pytorch toolbox, and all experiments are conducted on a single NVIDIA RTX 3090. We will release our code after the paper is accepted.

\subsection{Evaluation on Noisy FER Datasets}


In this section, we present a quantitative evaluation of the proposed ReSup method compared to other state-of-the-art approaches for addressing noisy labels in FER on RAF-DB, FERPlus, and Affectnet datasets. Following prior studies~\cite{wang2020suppressing,zhang2021relative,jiang2021boosting,she2021dive}, we randomly select a portion (10\%, 20\%, and 30\%) of the training data and corrupt their labels by assigning them to other random facial expression categories to generate noisy labels.


The experimental results in Table \ref{tab:noisy} demonstrate that the proposed ReSup method achieves superior performance compared to other methods. For instance, under the noise rate of 30\%, ReSup outperforms SCN by 9.4\%, 4.27\%, and 4.09\% on RAF-DB, FERPlus, and AffectNet, respectively. Moreover, the performance degradation of ReSup is only 2.29\% and 2.11\% when adding 30\% noise to original RAF-DB and FERPlus datasets, while SCN degrades by 9.57\% and 5.54\%, and PT drops by 4.37\% and 2.87\%.


It is noteworthy that ReSup uses a modeling-based approach for importance weight estimation, unlike SCN~\cite{wang2020suppressing}, DMUE~\cite{she2021dive}, and RUL~\cite{zhang2021relative} that rely on a network branch for this task. As such, ReSup does not face the concern that the powerful learning ability of deep neural networks (DNNs) may degrade the importance weight decision process. In comparison to PT~\cite{jiang2021boosting}, ReSup does not require knowledge of the exact noise level to set a threshold for selecting clean samples. Moreover, PT introduces a considerable amount of additional data ($280,000$ extra samples) for achieving better results through semi-supervised learning. Furthermore, SCN, RUL, and DMUE rely on a label correction module to improve performance, whereas ReSup outperforms them with only a more reliable utilization of the importance weights.

\subsection{Visualization Analysis}



In this section, we present a comparative analysis of the proposed ReSup method with other state-of-the-art FER schemes for noisy label data, namely RUL and SCN. We assess the effectiveness of these methods on the RAF-DB dataset with a 30\% noise level by visualizing and comparing the estimated importance scores. The results are showcased in Figure \ref{fig:weiexa}, where the first two columns depict clearly mislabeled images, and the last column features clean images. We observe that both the proposed ReSup method and RUL accurately distinguish between clean and noisy samples. In contrast, SCN may confuse clean and noisy samples. However, we note that RUL does not provide significant differences in uncertainty values for clean and noisy samples. Additionally, the image in the third column is ambiguous, and the proposed method assigns an importance weight close to $0.5$. However, SCN and RUL classify this image either as clean or noisy. Furthermore, the fourth column displays a meaningless image, and both RUL and ReSup assign small importance to it, while SCN does not.


Furthermore, we evaluated our ReSup on original FER datasets, which inevitably suffer from label noise, and the results are presented in section \ref{subsec:comsota}. We illustrated some samples with low importance weights from the original RAF-DB in Figure \ref{fig:weiraf}. We found that ReSup is more likely to assign low-importance weights to ambiguous, low-quality, and occluded images.

\begin{figure}
  \centering
   \includegraphics[width=0.85\linewidth]{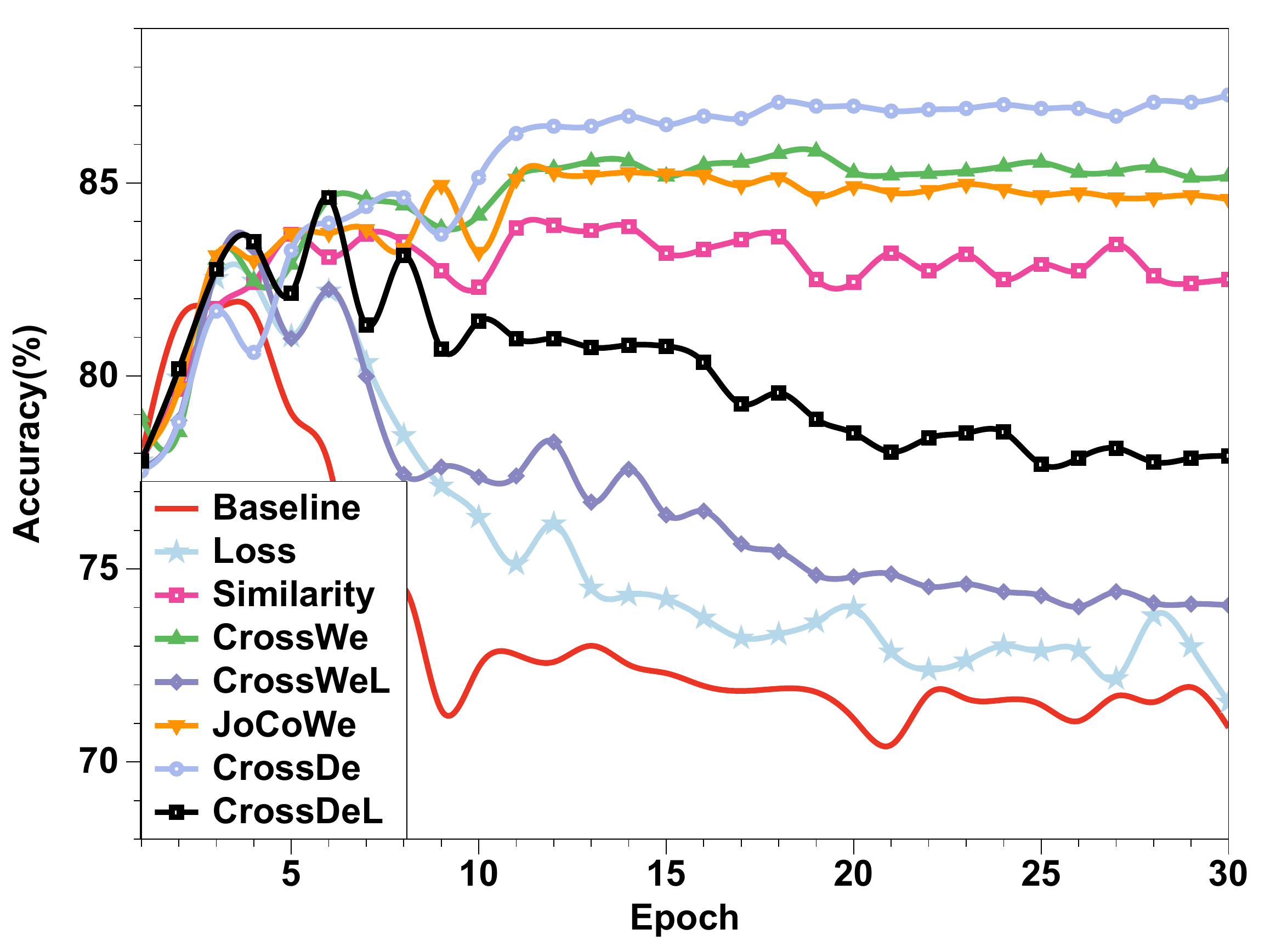}
   \caption{The test accuracy vs. epochs on RAF-DB with 30\% noise level. Loss and Sim represent using a single network to model and suppress label noise by relying on the loss and the similarity distribution, respectively. CrossDe is the proposed ReSup, while Crosswe is CrossDe w/o the consistency loss. CrossDeL and CrossWeL are similar to CrossDe and CrossWe but use the loss to model noise. JoCoWe is the JoCoR-based method.}
   \label{fig:ablraf}
\end{figure}

\begin{table}[htp]
  \centering
  \caption{Ablation studies on RAF-DB and FERPlus with 30\% noise. Loss and Sim represents loss and similarity-based noise modeling, respectively. WeEx and Col represents weight exchange and consistency loss, respectively.}
  \begin{tabular}{@{}lccccc@{}}
    \toprule
    Loss & Sim & WeEx & CoL & RAF-DB & FERPlus\\
    \midrule
     &  &  &  & 71.43\% & 72.14\% \\
    \checkmark &  &  &  & 72.67\%  & 78.23\%  \\
     & \checkmark &  &  & 83.05\% & 84.79\% \\
    \checkmark &  & \checkmark &  & 74.14\% & 80.83\% \\
     & \checkmark & \checkmark &  & 85.26\% & 85.74\% \\
    \checkmark &  & \checkmark & \checkmark & 77.91\% & 82.22\% \\
     & \checkmark & \checkmark & \checkmark & 86.86\% & 86.74\% \\
    \bottomrule
  \end{tabular}
  \label{tab:abl}
\end{table}

\subsection{Ablation Studies}
\label{subsec:abl}

To analyze the contribution of ReSup, we conduct ablation studies on RAF-DB and FERPlus with 30\% noise level. For implementing ReSup without the consistency loss, we set the $\lambda$ in equation \ref{equ:ljo} to $0$. Additionally, to further investigate the effectiveness of ReSup, we compare it with JoCoR~\cite{wei2020combating}, which selects examples based on the average loss of two networks. To simulate JoCoR, we compute the average importance weight of the two networks and use it to weigh the loss. We also compare our proposed label noise modeling approach with other methods~\cite{arazo2019unsupervised,li2020dividemix}, which model label noise by relying on the loss distribution. To implement these methods, we follow the approach outlined in~\cite{arazo2019unsupervised} and replace the similarity with the loss in our scheme.


\begin{figure}[t]
  \centering
   \includegraphics[width=0.85\linewidth]{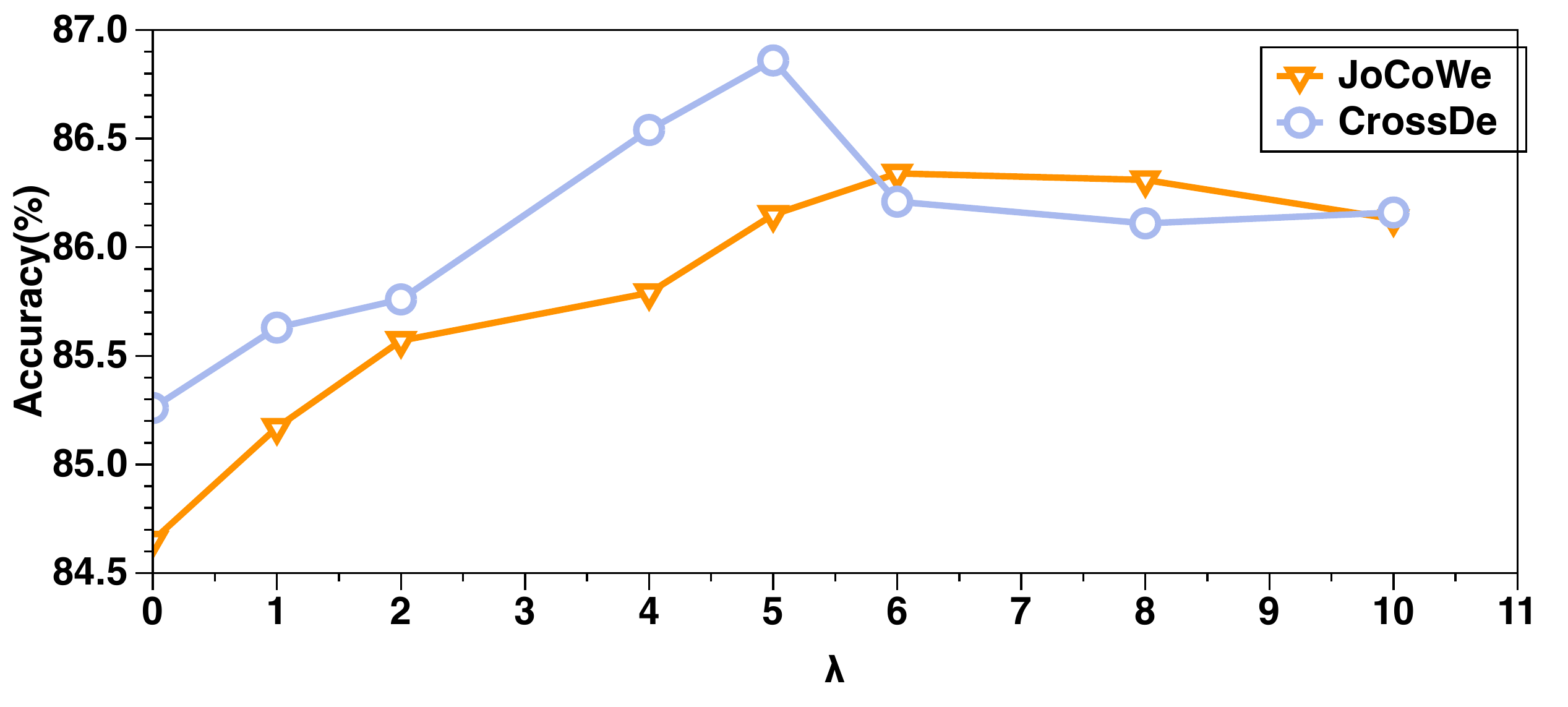}
   \caption{The accuracy of CrossDe and JoCoWe with different $\lambda$ on RAF-DB with 30\% noise. (For JoCoWe, the hyperparameter for the contrastive loss is $0.1*\lambda$)}
   \label{fig:hyp}
   \vspace{-0.1in}
\end{figure}

\begin{figure*}
  \centering
  \subfigure[The loss distribution of samples in the disgust class after the 1st epoch of training.]{
    \includegraphics[width=0.32\linewidth]{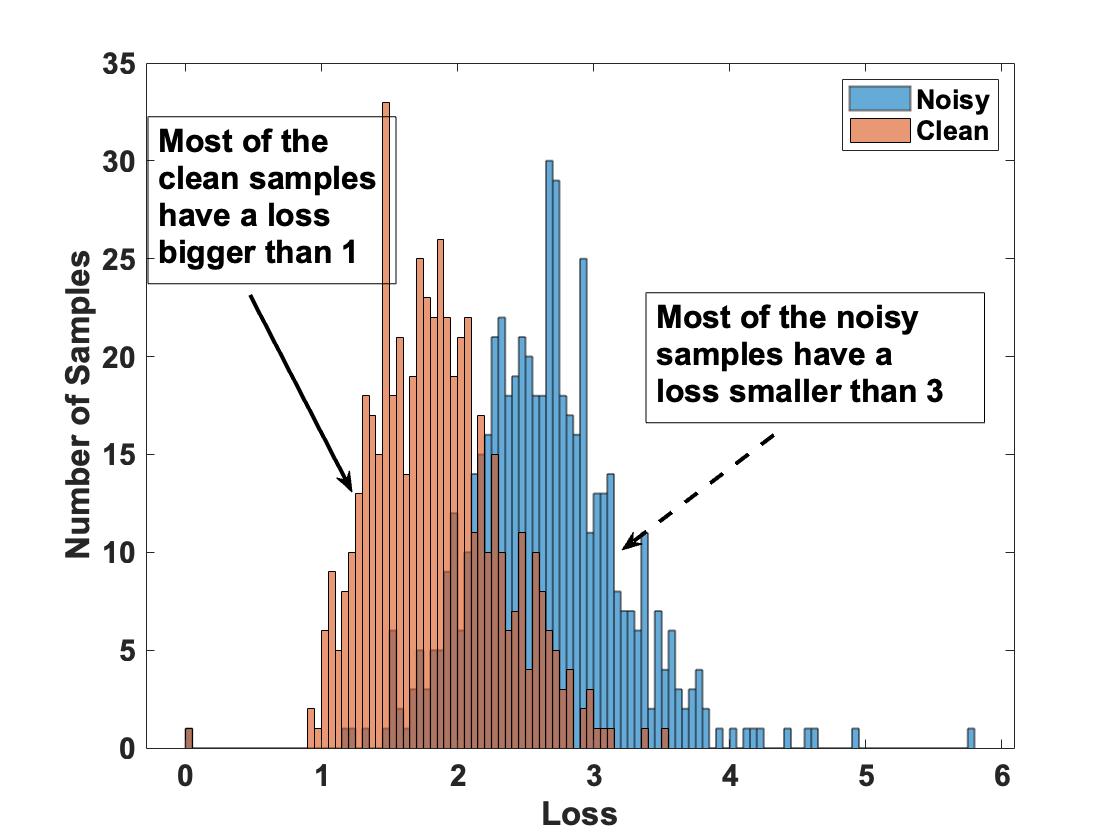}
    \label{fig:short-a}}
  \hfill
  \subfigure[The loss distribution of all samples after the 1st epoch of training.]{
    \includegraphics[width=0.32\linewidth]{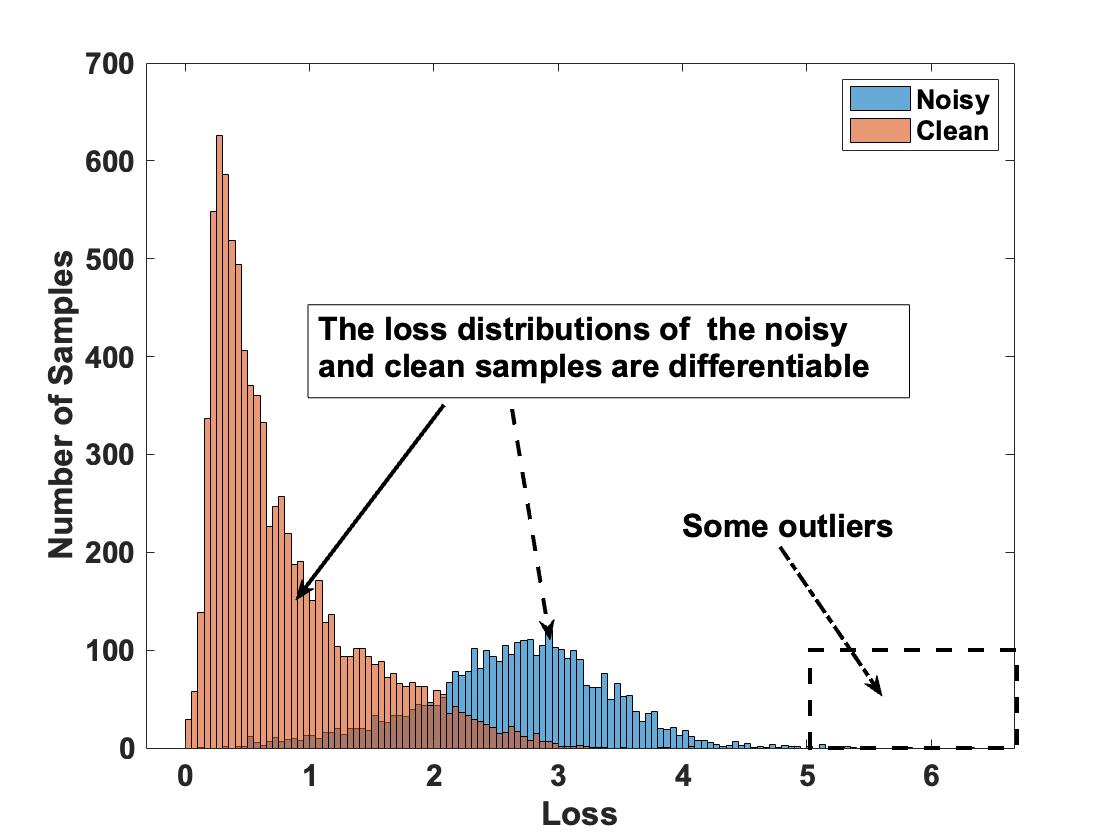}
    \label{fig:short-b}}
  \hfill
  \subfigure[The similarity distribution of all samples after the 1st epoch of training.]{
    \includegraphics[width=0.32\linewidth]{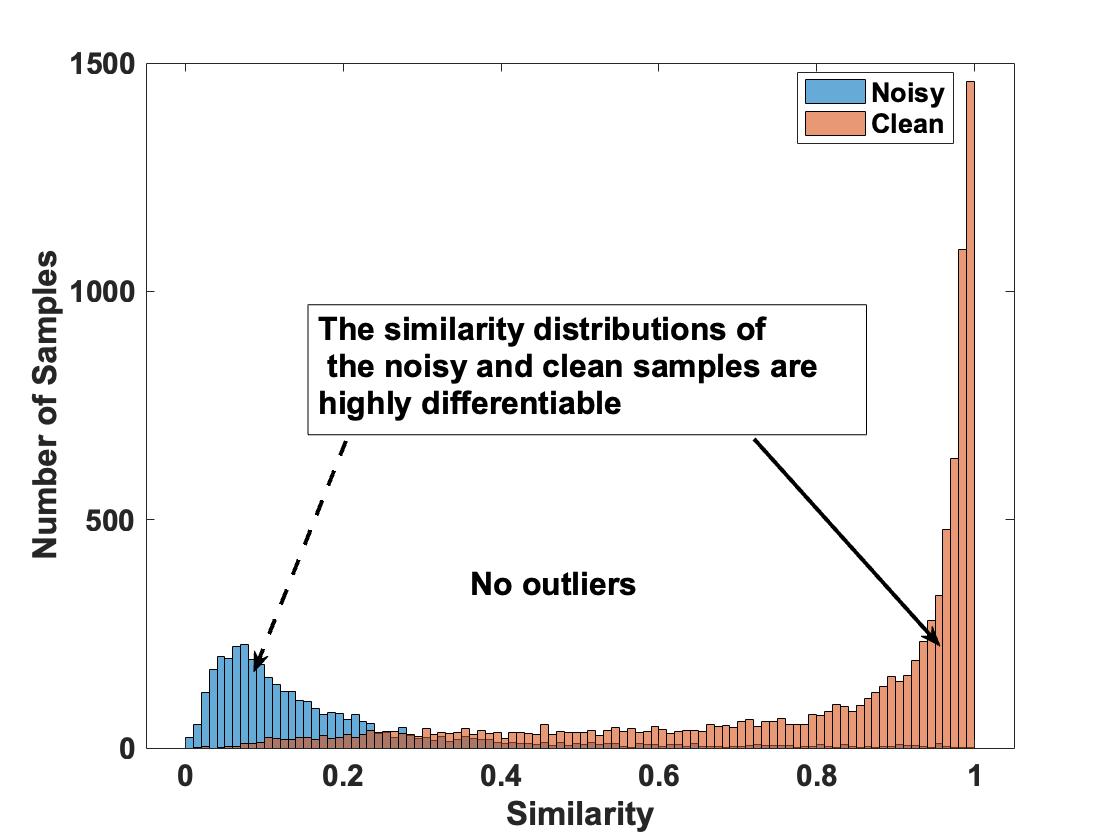}
    \label{fig:short-c}}

  \subfigure[The loss distribution of samples in the surprise class after the 1st epoch of training.]{
    \includegraphics[width=0.32\linewidth]{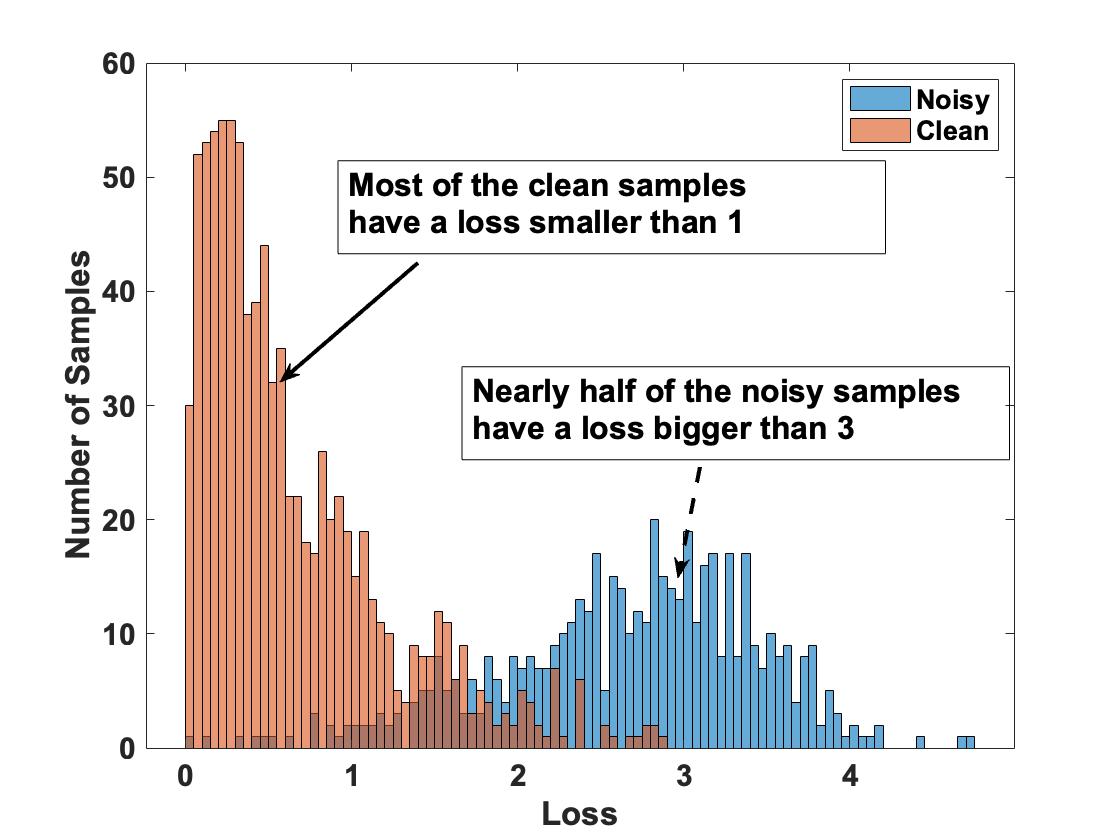}
    \label{fig:short-a1}}
  \hfill
  \subfigure[The loss distribution of all samples after the 7th epoch of training.]{
    \includegraphics[width=0.32\linewidth]{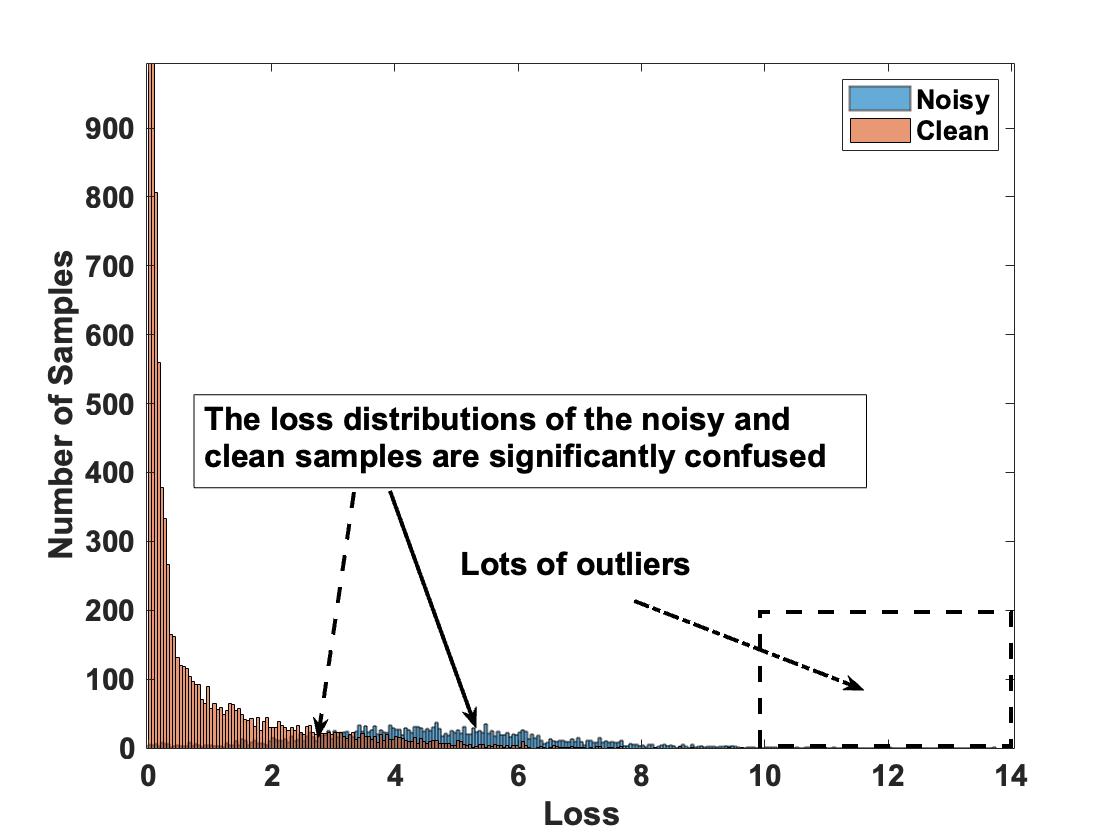}
    \label{fig:short-b1}}
  \hfill
  \subfigure[The similarity distribution of all samples after the 7th epoch of training.]{
    \includegraphics[width=0.32\linewidth]{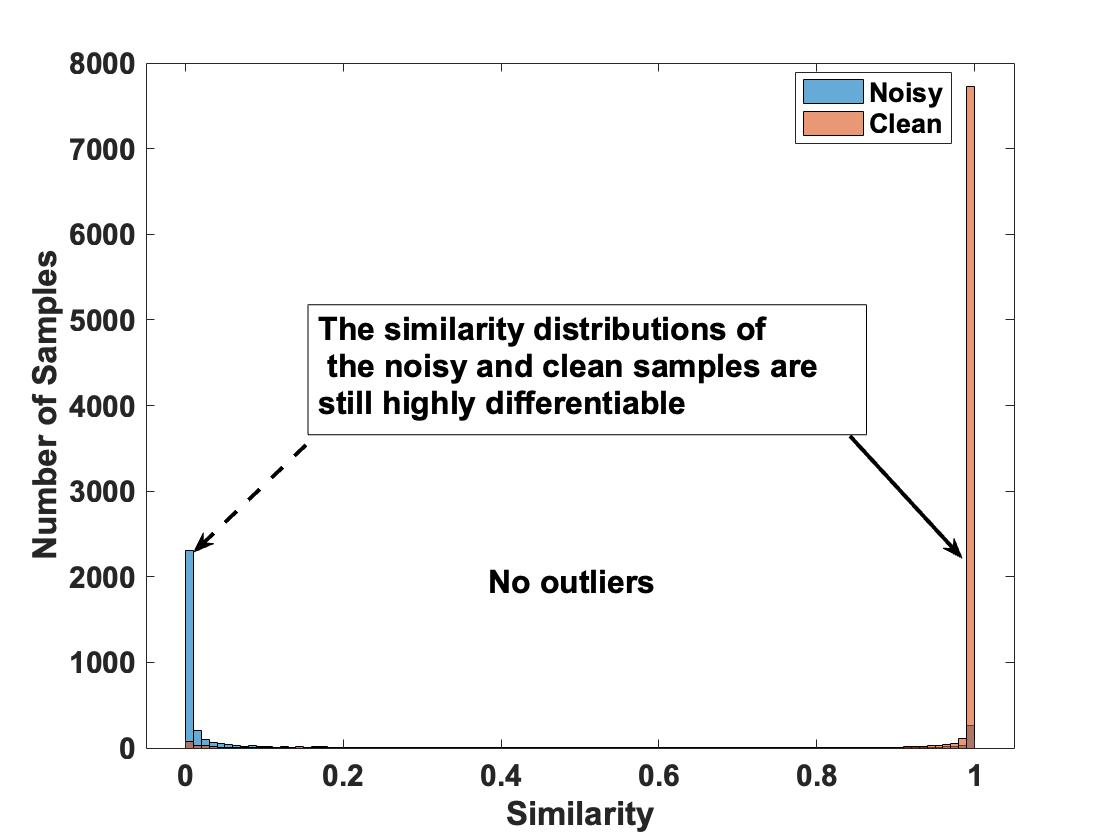}
    \label{fig:short-c1}}
  \caption{Loss and similarity distributions of training samples in the RAF-DB under 30\% noise.}
  \label{fig:short}
\end{figure*}


The experimental results are shown in Figure \ref{fig:ablraf} and Table \ref{tab:abl}, respectively. The observations are as follows: \textbf{(1)} The use of the loss for label noise modeling is found to be less effective as the training progresses, while the proposed similarity-based approach is seen to remain effective. \textbf{(2)} The weight exchange strategy is observed to improve performance by eliminating accumulated errors caused by unreliable weights. \textbf{(3)} The application of the consistency loss further enhances performance by preventing the networks from fitting unreliable weights when two networks' predictions significantly differ. \textbf{(4)} The proposed ReSup approach outperforms the JoCoR-based method due to the latter's clean sample selection method weak in eliminating accumulated errors.


We provide explanations as to why loss-based noise modeling may be ineffective. Firstly, these methods assume that the loss distributions of clean and noisy samples are identical across all classes. However, in the case of FER, inter-class similarity and ambiguity result in varying learning difficulty for different classes, leading to diverse loss distributions across different categories (Figures \ref{fig:short-a} and \ref{fig:short-a1}). Consequently, the interval of the distribution of all noise samples is large (Figure \ref{fig:short-b}). Secondly, during noise-robust training, noisy samples are often ignored to prevent the model from fitting to them, leading to decreased prediction probability $p$ as model performance improves. However, CE loss is sensitive to small $p$ ($loss = -log(p)$), resulting in a significant variation in loss values with slight changes in $p$. This sensitivity amplifies the difference in the loss distribution of noisy samples from different classes, causing the interval of the distribution to become larger, and the approximately uniform distribution of all noisy samples (Figure \ref{fig:short-b1}). Consequently, it becomes difficult to model the distributions of clean and noisy samples using a mixture model, which is unsupervised. The proposed similarity-based approach, which overcomes the limitations caused by CE loss, is demonstrated in Figures \ref{fig:short-c} and \ref{fig:short-c1}. Further details can be found in the supplementary materials.



We evaluate the effect of the hyperparameter $\lambda$ for both ReSup and JoCoR-based methods as shown in Figure \ref{fig:hyp}. ReSup achieved the best performance at $\lambda = 5$, while ReSup performs better than JoCoR-based method with different $\lambda$. 


\begin{table}
  \centering
  \caption{Comparison with other state-of-the-art FER schemes.}
  \begin{tabular}{@{}lccc@{}}
    \toprule
    Method & RAF-DB & FERPlus & AffectNet \\
    \midrule
    gACNN~\cite{li2018occlusion} & 85.07\% & - & 58.78\%\\
    IPA2LT~\cite{zeng2018facial} & 86.77\% & - & 55.71\%\\
    LDL-ALSG~\cite{chen2020label} & 85.53\% & - & 59.35\%\\
    EfficientFace~\cite{zhao2021robust} & 88.36\% & - & 63.70\%\\
    SCN~\cite{wang2020suppressing} & 87.03\% & 88.01\% & - \\
    DMUE~\cite{she2021dive} & 88.76\% & 88.64\% & - \\
    RUL~\cite{zhang2021relative} & 88.98\% & 88.75\% & -\\
    PT~\cite{jiang2021boosting} & 88.69\% & 86.60\% & 58.54\%\\
    Ours & \textbf{89.70\%} & \textbf{88.85\%} & \textbf{65.46\%}\\
    \bottomrule
  \end{tabular}
  \label{tab:sotacom}
  \vspace{-0.15in}
\end{table}

\subsection{Comparision with State-of-the-art Methods}
\label{subsec:comsota}


We compare ReSup with several state-of-the-art FER methods on original RAF-DB, FERPlus, and AffectNet in Table \ref{tab:sotacom}. Besides the noisy label FER studies mentioned in previous sections, LDL-ALSG~\cite{chen2020label} uses topological information from relevant tasks to construct label distributions to guide the training of the model. EfficientFace~\cite{zhao2021robust} incorporates local and global features to learn a FER model. IPA2LT~\cite{zeng2018facial} introduces the latent ground truth for learning with inconsistent annotations. gACNN~\cite{li2018occlusion} leverage attention mechanism to conduct an occlusion-aware FER model. Our method outperforms these recent state-of-the-art methods with 89.70\%, 88.85\%, and 65.46\% on RAF-DB, FERPlus, and AffectNet, respectively. 


We also implement two plain methods using threshold from Co-teaching~\cite{han2018co} and PT~\cite{jiang2021boosting} with performance of (86.53\%,83.21\%,82.37\%) and (84.91\%,77.93\%,72.46\%) on the RAF-DB (noise 10\%-30\%), respectively. We also meticulously tune the threshold-based method from~\cite{han2018co,jiang2021boosting} on our scheme, with the best performance of 87.32\%, 84.68\%, and 83.51\%. The performance of ReSup exceeds all these solutions. Our ReSup outperforms all of these solutions and offers an advantage over them by not requiring prior knowledge or estimation of the noise ratio to carefully tune the parameters for optimal performance.


\subsection{Experiments on Real Noisy FER Dataset}
\label{subsec:expreno}


To validate the efficacy of ReSup, we experiment on a real-world noisy FER dataset, ExpW~\cite{zhang2018facial}, which comprises a large number of low-quality annotations. We train our models on this dataset and evaluate on the ExpW/RAF-DB test sets. The results show that our ReSup (73.12\%/75.65\%) is more effective compared with the baseline (67.87\%/71.77\%).

\begin{table}[htp]
  \centering
  \caption{The influence of different backbones on ReSup. We carry out experiments on RAF-DB and $^*$ means baseline.}
  \begin{tabular}{@{}lcccc@{}}
    \toprule
    Noise & MobileNet & ResNet18 & ResNet50 & VGG16 \\
    \midrule
    20\%$^*$ & 74.54\% & 74.25\% & 76.01\% & 72.49\% \\
    20\% & 84.49\% & 84.55\% & 86.86\% & 83.28\% \\
    30\%$^*$ & 68.52\% & 68.19\% & 68.87\% & 66.33\% \\
    30\% & 83.57\% & 82.76\% & 85.24\% & 82.50\% \\
    \bottomrule
  \end{tabular}
  \label{tab:difnet}
\end{table}

\subsection{Experiments on Different Network Structures}
\label{subsec:expdin}

We have conducted experiments on different network structures as shown in Table \ref{tab:difnet}, and all networks are solely pre-trained on ImageNet. Our scheme is effective on various network structures.

\subsection{Experiments on CIFAR-10}
\label{subsec:expci}


We evaluated the generalization ability of ReSup on CIFAR10, where ResNet18 was replaced with the 7-layer CNN utilized in JoCoR~\cite{wei2020combating} for a fair comparison. Experimental results are presented in Table \ref{tab:cifarln}. Our approach outperformed JoCoR and Jo-SRC~\cite{yao2021jo}, demonstrating its effectiveness. Further experimental analysis on CIFAR10 and CIFAR100 is included in the supplementary materials.

\begin{table}[htp]
  \centering
  \caption{Average test accuracy on CIFAR-10 over the last 10 epochs, and $^\ddag$ means asym noise.}
  \begin{tabular}{@{}lcccc@{}}
    \toprule
    Noise & 20\% & 50\% & 80\% & 40\%$^\ddag$ \\
    \midrule
    Standard & 69.18\% & 42.71\% & 16.24\% & 69.43\% \\
    JoCoR~\cite{wei2020combating} & 85.73\% & 79.41\% & 27.78\% & 73.78\% \\
    Jo-SRC~\cite{yao2021jo} & 87.80\% & 73.67\% & 34.30\% & 79.96\% \\
    ReSup & \textbf{90.08\%} & \textbf{86.06\%} & \textbf{70.78\%} & \textbf{85.31\%} \\
    \bottomrule
  \end{tabular}
  \label{tab:cifarln}
\end{table}

\section{Conclusion}


In this paper, we have proposed ReSup, a novel approach for tackling the problem of noisy label FER. Our method consists of two key components: label noise modeling and noise-robust learning. The former enables us to reliably model the label noise in FER, while the latter allows us to mitigate the negative impact caused by unreliable weights and learn from noisy datasets. Extensive experiments on three public datasets have demonstrated the effectiveness of our approach, which outperforms several state-of-the-art FER methods. Our approach is also shown to be effective on various network structures and able to generalize well to other datasets.  Overall, our work provides a promising solution for improving the performance of FER models in the presence of noisy labels.


{\small
\bibliographystyle{ieee_fullname}
\bibliography{egbib}
}

\end{document}